\documentclass{article}
\usepackage{spconf,amsmath,graphicx,url,multirow}
\usepackage{bm,multirow,tabularx, threeparttable, amssymb}
\def\x{{\mathbf x}}

\def\x{{\bm x}}

\def\l{{\bm l}}
\def\bpi{{\bm \pi}}
\def\B{{\mathcal B}}
\def\bt{{\bm \theta}}

\title{CAT: CRF-based ASR Toolkit}
\name{Keyu An, Hongyu Xiang, Zhijian Ou$^{\dagger}$\thanks{This work is supported by NSFC 61976122. $\dagger$ Corresponding author.}}\address{Speech Processing and Machine Intelligence (SPMI) Lab, Tsinghua University, China\\aky19@mails.tsinghua.edu.cn, xianghy16@mails.tsinghua.edu.cn, ozj@tsinghua.edu.cn}

\begin{document}
\ninept
\maketitle
\begin{abstract}
In this paper, we present a new open source toolkit for automatic speech recognition (ASR), named CAT (CRF-based ASR Toolkit). A key feature of CAT is discriminative training in the framework of conditional random field (CRF), particularly with connectionist temporal classification (CTC) inspired state topology. CAT contains a full-fledged implementation of CTC-CRF and provides a complete workflow for CRF-based end-to-end speech recognition. Evaluation results on Chinese and English benchmarks such as Switchboard and Aishell show that CAT obtains the state-of-the-art results among existing end-to-end models with less parameters, and is competitive compared with the hybrid DNN-HMM models. 
Towards flexibility, we show that i-vector based speaker-adapted recognition and latency control mechanism can be explored easily and effectively in CAT.
We hope CAT, especially the CRF-based framework and software, will be of broad interest to the community, and can be further explored and improved.
 
\end{abstract} 
\section{Introduction}
\label{sec:intro}
In addition to theories and algorithms, open source  toolkits make substantial contributions to automatic speech recognition (ASR) technologies.  
A good ASR toolkit is an integration of good algorithms, efficient implementations and manageable code bases. 
In recent years, significant advancement has been made by the application of deep neural networks (DNNs). Recent toolkits using DNNs such as 
Kaldi \cite{Povey2012KALDI}, 
Eesen \cite{Miao2015EESEN}, 
Wav2letter++ \cite{Pratap2018wav2letter} and 
ESPnet \cite{Watanabe2018ESPnet} promote the development and application of ASR technologies, with varying pipelines and concerns. Toolkits based on DNN-HMM hybrid systems like Kaldi \cite{Povey2012KALDI} and RASR \cite{RASR} achieve the state-of-the-art performance in terms of recognition accuracy, usually measured by word error rate (WER) or character error rate (CER). In contrast, end-to-end systems\footnote{We follow the definition of end-to-end in \cite{SS-LF-MMI}: ``flat-start training of a single DNN in one stage without using any previously trained models, forced alignments, or building state-tying decision trees.''} put simplicity of the training pipeline at a higher priority and usually are data-hungry.

In this paper we present CAT (CRF-based ASR Toolkit)\footnote{\url{https://github.com/thu-spmi/cat}}, which aims at combining the advantages of the two kinds of systems. CAT advocates discriminative training in the framework of conditional random field (CRF), particularly with but not limited to connectionist temporal classification (CTC) \cite{graves2006connectionist} inspired state topology.
The recently developed CTC-CRF (namely CRF with CTC topology) \cite{Xiang2019CRF} has achieved superior benchmarking performance with training data ranging from $\sim$100 to $\sim$1000 hours, while being end-to-end with simplified pipeline and being data-efficient in the sense that cheaply available language models (LMs) can be leveraged effectively with or without a pronunciation lexicon.

Major features of CAT are as follows.

\textbf{1.} 
CAT contains a full-fledged implementation of CTC-CRF. 
A non-trivial issue is that the gradient in training CRFs is the difference between empirical expectation and model expectation, which both can be efficiently calculated by the forward-backward algorithm
\footnote{Calculating the two expectations is similar to calculations over the numerator graph and denominator graph in LF-MMI respectively \cite{povey2016purely}.}.
CAT modifies warp-ctc \cite{warp-ctc} for fast parallel calculation of the empirical expectation, which resembles the CTC forward-backward calculation  \cite{graves2006connectionist}.
CAT calculates the model expectation using CUDA C/C++ interface, drawing inspiration from Kaldi's implementation of denominator forward-backward calculation.

\textbf{2.} 
CAT adopts PyTorch \cite{paszke2017automatic} to build DNNs and do automatic gradient computation, and so inherits the power of PyTorch in handling DNNs.

\textbf{3.} 
CAT provides a complete workflow for CRF-based end-to-end speech recognition.
CAT provides complete training and testing scripts for a number of Chinese and English benchmarks and all the experimental results reported in this paper can be readily reproduced. Detailed documentation and code comments are also provided in CAT, making it easy to get start and obtain state-of-the-art baseline results even for beginners of ASR.

\textbf{4.} 
Evaluation results on major benchmarks such as Switchboard and Aishell show that 
CAT obtains the state-of-the-art results among existing end-to-end models with less parameters, and is competitive compared with the hybrid DNN-HMM models.

\textbf{5.}
Towards flexibility, we show that i-vector based speaker-adapted recognition and latency control mechanism can be explored easily and effectively in CAT.

\section{Related work}
\label{sec:format}
Currently, speech recognition models can be roughly divided into two categories: DNN-HMM hybrid models and end-to-end models. 
One of the feature of Kaldi is its efficient implementation of the lattice-free maximum-mutual-information (LF-MMI) model, which is a typical hybrid DNN-HMM model.
The pipeline consists of initial GMM-HMM training, and iterative context tree building and forced alignment. 
End-to-end LF-MMI \cite{SS-LF-MMI} (EE-LF-MMI) has been developed, with two versions of using mono-phones or tree-free bi-phones.
The differences between EE-LF-MMI and CTC-CRF are detailed in \cite{Xiang2019CRF}.
It is shown in our experiments that CTC-CRF mono-phone system matches EE-LF-MMI bi-phone system.
 
End-to-end models aim to directly map the speech sequence (raw audio, spectrum features, etc.) to the label sequence (words, phonemes, etc.) with minimum intermediate components.
Three main classes of end-to-end models are based on CTC \cite{graves2006connectionist}, attention based Seq2Seq \cite{zeyer2018improved} and RNN-transducer (RNN-T)  \cite{graves2012sequence} respectively.
As representative toolkits, Eesen is based on regular CTC, and ESPnet relies on attention and adopts hybrid CTC/attention.
Wav2letter++ is known for its efficiency with fast tensor operations and the use of pure convolutional neural networks in both acoustic and language models.

End-to-end models have received increasing interests and achieved performances close to the hybrid models on a few benchmarks, but still faces a number of issues. First, there is still a pronounced gap between attention end-to-end models and hybrid models in terms of recognition accuracy \cite{vs}. Second, the recognition accuracy of the hybrid models can be further boosted with classical speech recognition techniques such as i-vector based speaker-adapted recognition \cite{Saon2014Speaker}, while the application of these techniques to end-to-end systems have not been thoroughly explored.
A third issue sometimes overlooked for end-to-end ASR toolkits is the demand for low latency recognition which is crucial for streaming ASR applications.
CTC-based Eesen uses bidirectional models by default.
In attention based end-to-end systems (e.g. ESPnet), bidirectional encoder and global soft attention present inherent difficulty for low latency.
There are recent efforts such as using monotonic chunkwise attention (MoChA) \cite{MoChA}, Latency-controlled BLSTM \cite{LC-BLSTM}.
Online recognition with ESPnet has been recently studied \cite{is19_yan}.
Wav2letter++ is based solely on convolutional neural networks, which use restricted future context and realize low latency. However, in order to model long-range dependencies, the neural network in \cite{Zeghidour2018Fully} is extremely deep and big (with 100 million parameters).
Remarkably, time-delay neural networks (TDNNs) with interleaving LSTM layers (TDNN-LSTM) \cite{Low-latency} is used in Kaldi to limit the latency while maintain the recognition accuracy. 

Finally, for programming languages used in toolkits, Kaldi core primarily uses C++ which is efficient but not flexible in supporting various rapid developments in DNNs.
PyTorch-Kaldi \cite{Ravanelli2018THE} builds the neural networks with PyTorch, and PyKaldi \cite{Pykaldi} allows users to interact with Kaldi and OpenFst via Python language.

\section{CRF-based ASR}
\label{sec:Discriminative training with CRF}
CAT conducts discriminative training of acoustic model (AM) based on conditional maximum likelihood  \cite{Xiang2019CRF}:
\begin{equation} \label{eq:crf-obj1}
\mathcal{J}_{CRF}(\bt) = \log p_{\bt}(\l|\x)
\end{equation}
where $\x \triangleq x_1,\cdots\, x_T$ is the speech feature sequence and $\l \triangleq l_1, \cdots\, l_L$ is the label sequence.
The label could be phone, character, word-piece and etc.
Note that $\x$ and $\l$ are in different lengths and usually not aligned in training data.
The alignment could be handled implicitly by attention or explicitly by introducing a hidden state sequence $\bpi \triangleq \pi_1,\cdots\,\pi_T$.
CRF-based ASR takes the later approach, which is also taken in DNN-HMM hybrid, CTC and RNN-T.
When introducing $\bpi$,
state topology refers to the state transition structure in $\bpi$, which basically 
defines a mapping $\mathcal{B}: S_\pi^{*} \to S_l^{*}$ that maps a state sequence $\bpi$ to a unique label sequence $\l$.
Here $S_l^{*}$ denote the set of all sequences over the alphabet $S_l$ of labels, and $S_\pi^{*}$ similarly for the alphabet $S_\pi$ of states.
It can be seen that HMM, CTC, and RNN-T implement different topologies\footnote{
These topologies can all be used in CRF-based ASR, which would be our future work. CAT toolkit currently supports CRF with CTC topology.}. 
CTC topology defines a mapping that removes consecutive repetitive labels and blanks, with $S_\pi$ defined by adding a special blank symbol $<$blk$>$ to $S_l$.
CTC topology is appealing, since it allows a minimum size of $S_\pi$ and avoids the inclusion of silence symbol, as discussed in \cite{Xiang2019CRF}.

Then the posteriori of $\l$ can be defined through the posteriori of $\bpi$ as follows:
\begin{equation} \label{eq:post-l} 
	p_{\bt}(\l | \x) = \sum_{\bpi \in \mathcal{B}^{-1}(\l)} p_{\bt}(\bpi | \x)
\end{equation}
And the posteriori of $\bpi$ can be further defined by a CRF:
\begin{equation} \label{eq:post-pi}
p_{\bt}(\bpi|\x) = \frac{\exp(\phi_{\bt}(\bpi, \x))}{\sum_{\bpi'}{\exp(\phi_{\bt}({\bpi', \x}))}}
\end{equation}
Here $\phi_{\bt}(\bpi, \x)$ denotes the potential function of the CRF, defined as:
\begin{displaymath}
\phi_{\bt}(\bpi, \x) = \log p(\l)+ \sum_{t=1}^{T} \log p_{\bt}(\pi_t|\x)
\end{displaymath}
where $\l = \B(\bpi)$. 
$\sum_{t=1}^{T} \log p_{\bt}(\pi_t|\x)$ defines the node potential, calculated from the bottom DNN.
$\log p(\l)$ defines the edge potential, often realized by an n-gram LM of labels and, for reasons to be clear in the following, referred to as the denominator n-gram LM.
Remarkably, regular CTC suffers from the conditional independence between the states in $\bpi$. In contrast, by incorporating $\log p(\l)$ into the potential function in CRF-based ASR, this drawback is naturally avoided.
Combing Eq. (\ref{eq:crf-obj1})-(\ref{eq:post-pi}) yields the CRF objective function specifically as:
\begin{equation} \label{eq:crf-obj2}
\mathcal{J}_{CRF}(\bt) = \log \frac{  \sum_{\bpi \in \mathcal{B}^{-1}(\l)} \exp(\phi_{\bt}(\bpi, \x))}{\sum_{\bpi'}{\exp(\phi_{\bt}({\bpi', \x}))}}
\end{equation}

The gradient of the above objective function involves  two gradients calculated from the numerator and denominator respectively, which essentially correspond to the two terms of empirical expectation and model expectation as commonly found in estimating CRFs.
Similarly to LF-MMI, both terms can be efficiently obtained via the forward-backward algorithm, as further detailed in Section \ref{sec:numerator/denominator calculation}.
Especially, the denominator calculation involves running the forward-backward algorithm over the denominator graph $\bf{T}_{den}$, represented as a weighted finite sate transducer (WFST).
$\bf{T}_{den}$ is an composition of the CTC topology WFST and the WFST representation of the n-gram LM of labels, which is called the denominator n-gram LM, to be differentiated from the word-level LM in decoding.

\section{Implementation}
\label{sec:Implementation}

CAT consists of separable AM and LM, which meets our rationale to be data-efficient.
The AM training workflow in CAT is shown in the Fig 1. 
CAT uses SRILM for LM training, and some code from Eesen for decoding graph compiling and WFST based decoding.

\subsection{Kaldi style feature extraction}
\label{Kaldi style feature extraction}
CAT integrates Kaldi into the data preparation and feature extraction steps. Kaldi serves as a good reference, with complete modules for building ASR systems, covering steps from feature extraction to decoding and evaluation. 
Besides, the code organization of CAT also follows the Kaldi manner, which means 1) high-level workflows are expressed in shell scripts, and 2) the source code and the examples are separated, which facilitates the reuse of source code between exemplar systems over different speech datasets.

\begin{figure}[t]
	\centering
	\centerline{\includegraphics[width=9cm]{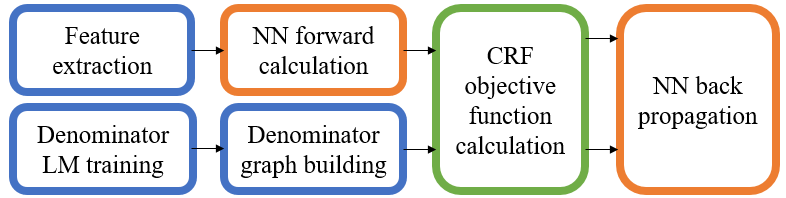}}
	\caption{The AM training pipeline in CAT. The blue block stands for Kaldi style data preparation, the orange for neural network training with PyTorch, and the green for our C++ implementation of CRF objective function calculation.}
	\label{fig:training pipeline}
\end{figure}

\subsection{PyTorch in acoustic modeling}
\label{PyTorch in acoustic modeling}
In acoustic modeling, we use PyTorch to build various neural networks with Python language. PyTorch supports dynamic neural network building, facilitating the exploration of complex neural network architectures. We provide the implementation of commonly used neural networks in ASR, e.g. LSTM with its variants and TDNN-LSTM. A wide variety of new elements in deep learning applications such as dropout \cite{Srivastava2014Dropout} is available in PyTorch, with which we can easily incorporate those new elements in building ASR systems. 
Besides, PyTorch provides support for multiprocess parallelism on one or more machines, enabling us to conduct experiments with large-scale training data. 

\subsection{Numerator and Denominator calculation}
\label{sec:numerator/denominator calculation}
The numerator calculation is similar to the regular CTC computation. We use warp-ctc \cite{warp-ctc} to calculate the numerator objective function in Eq. (\ref{eq:crf-obj2}). The input of the warp-ctc is modified to be the log-softmax output of the bottom neural network, rather than the linear layer output.
Note that for label sequence $\l$,  $\log p(\l)$ also appears in the numerator but behaves like an constant. So $\log p(\l)$ is pre-calculated based on the denominator n-gram LM and saved, and then applied in training.

The denominator calculation is implemented via CUDA C/C++ on GPUs, drawing inspiration from LF-MMI in Kaldi. The main differences are that 1) our implementation is in the log domain, which is more stable than in the original numeric domain as in Kaldi. 2) Varying lengths of utterances are supported, rather than modifying the utterance length to one of 30 lengths; and we do not use the leaky-HMM trick.

\subsection{WFST-based decoding}
\label{WFST for decoding}
We adopt the WFST-based decoding in Eesen \cite{Miao2015EESEN}. 
Notably, as \cite{Xiang2019CRF} pointed out, the CTC topology WFST in Eesen (denoted by T.fst) is not correct with the inappropriate use of $<$blk$>$ symbol, making the decoding graph mistakenly larger. The mistake is corrected in CAT.

As mentioned before, CAT adopts CTC topology instead of HMM topology, which benefits the decoding process in two aspects. First, the size of the symbol alphabet $S_\pi$ of CTC topology, which equals to the size of $S_l$ plus one (for $<$blk$>$), is far less than the alphabet size of HMM topology used in \cite{SS-LF-MMI}. Second, as figured out in \cite{graves2006connectionist}, the output of CTC is typically a series of spikes, separated by lots of ‘blanks’. Ignoring these frames with high blank scores does not affect beam decoding much, but the time cost will be reduced significantly. It is found in our experiments that ignoring frames with blank scores greater than 0.7 leads to 0.1\% reduction in recognition accuracy on Switchboard, but the decoding time is 30\% less than not ignoring. We also explore the application of the RNN language model in lattice rescoring.

\section{Experiment Settings}
The experiment consists of two parts. In the first part, we introduce the results on major benchmarks using state-of-the-art bidirectional recurrent networks. The second part presents our exploration on low-latency unidirectional models for streaming ASR.

\subsection{Setup for benchmarking experiment}
We compare the performance of CAT with state-of-the-art ASR systems on three open source benchmarks: 80hr WSJ, 170hr Aishell and 300hr Switchboard. Speed perturbation is used on all datasets to augment the training data. Unless otherwise stated, 40 dimension filter bank with delta and delta-delta features are used as input. The features are normalized via mean subtraction and variance normalization per utterance, and sampled by a factor of 3. 

Unless otherwise stated, the acoustic model is two blocks of VGG layers followed by a 6-layer bidirectional LSTM (BLSTM) similar to \cite{Hori2017Advances}. We apply 1D max-pooling to the feature maps produced by VGG blocks on the frequency dimension only, as the input features have been sampled in time-domain and we found that max-pooling along time dimension will deteriorate the recognition accuracy. The first VGG block has 3 input channels corresponding to spectral features, delta, and delta delta features. The BLSTM has 320 hidden units per direction for each layer and the total number of parameters is 16M, much smaller than most end-to-end models. For the training, a dropout \cite{Srivastava2014Dropout} probability of 50\% is applied to the LSTM layer to prevent overfitting.  Following \cite{Xiang2019CRF}, a CTC loss with a weight $\alpha$ is combined with the CRF loss to help convergence. We set $\alpha$ = 0.1 by default and we found in practice that the smaller $\alpha$ is, the better the final result will be, but the convergence time will be a bit longer.

On Switchboard we also evaluate our model with speaker adaptation \cite{Saon2014Speaker} and RNNLM lattice rescoring \cite{RNNrescore}. Speaker adaptation is an important component in state-of-the-art hybrid DNN-HMM recognizers, but end-to-end models tend to exclude speaker adaptation techniques to maintain the end-to-end training manner. Here we offer promising results of improvements from speaker adaptation techniques on end-to-end acoustic model. Specifically, we extract 100-dimensional i-vectors from MFCC features as in \cite{Peddinti2015ATD}. 
Thus in Switchboard experiments, the 120-dimensional MFCCs+$\Delta$+$\Delta \Delta$ appended with the i-vector are used as the input for neural networks.
We use MFCC features here to stay consistent with the i-vector extraction. The acoustic model is a 6-layer BLSTM with 220 input units. The VGG blocks are not used here.
\subsection{Setup for latency control experiment}
 The key to achieving high-accuracy online speech recognition is to exploit limited but sufficient future information with appropriate mechanisms. For example, the model in \cite{onlineNey} enables online-recognition by moving a window over the input stream of the bidirectional RNN, while \cite{warp-ctc} applies a modification to unidirectional RNN by employing a special layer named row convolution to provide a small portion of future information. We explore two methods for low latency. The first method is the use of VGG blocks mentioned previously, as the convolution layer can provide certain amount of future information. The second method is to use an structure similar to TDNN-LSTM proposed in \cite{Low-latency}. Results for standard unidirectional and bidirectional LSTM with the same number of layers  and hidden units per layer (as in VGG-LSTM) are also provided for comparison. Offline cepstral mean and variance normalization (CMVN) is not used in online experiment because it requires a full sentence input to extract the mean and variance information. Instead, we use batch normalization to speed up convergence.
\section{Experimental results}
\subsection{Results on  major benchmarks}
The WER results on WSJ are shown in Table 1. The evaluation dataset contains
the dev93 and eval92. On eval92, our performance is comparable with hybrid LF-MMI and flat-start EE-LF-MMI, and better than all other end-to-end models. On the more difficult dev93 dataset, CAT is only slightly worse ($\sim$ 7\%) than hybrid LF-MMI which uses bi-phone context dependency modeling.

\begin{table}[tbp]
	\centering
	\scalebox{0.88}{
	\begin{threeparttable}
	\begin{tabular}{|ccc|cc|}
		\hline
		Platforms    & Unit     & LM  & dev93 & eval92 \\
		\hline
	    EE-LF-MMI \cite{SS-LF-MMI} & mono-phone & 4-gram  & 6.3\% & 3.1\% \\
	    EE-LF-MMI \cite{SS-LF-MMI}& bi-phone & 4-gram  & 6.0\% & 3.0\% \\
	    LF-MMI \cite{SS-LF-MMI} & mono-phone & 4-gram  & 6.0\% & 3.0\% \\
        LF-MMI \cite{SS-LF-MMI} & bi-phone & 4-gram  & 5.3\% & 2.7\% \\		
		Eesen \cite{Miao2015EESEN} & mono-phone & 3-gram  & 10.87\% & 7.28\% \\

		ESPnet \cite{Watanabe2018ESPnet} & mono-char & LSTM   & 12.4\% & 8.9\% \\
        Wav2letter++ \cite{Zeghidour2018Fully} & mono-char & 4-gram  & 9.5\% & 5.6\% \\
        Wav2letter++   \cite{Zeghidour2018Fully}& mono-char & ConvLM  & 6.8\% & 3.5\% \\
		\hline
		CAT & mono-phone & 4-gram  & 5.7\% & 3.2\% \\
		\hline
	\end{tabular}
	\end{threeparttable}}
	\vspace{-0.1cm}
	\caption{WSJ results}
	\label{tab:wsj}
	\vspace{0.2cm}
\end{table}

The results on  Aishell are shown in Table 2. We use CER to evaluate the performance on mandarin benchmark by convention. We do not use pitch feature as in \cite{Povey2012KALDI} \cite{Watanabe2018ESPnet} \cite{ACSAttention}, because the pitch feature is not suitable for composing a 3-channel feature map together with the fbank feature. The result shows that CAT obtains state-of-the-art performance on Aishell dataset, the CER is much better than other end-to-end models and even the hybrid Kaldi-chain model. 

\begin{table}[tbp]
	\centering
	  \scalebox{0.88}{
	  \begin{threeparttable}
	  \begin{tabular}{|ccc|c|}
	  \hline
	  Model    & Unit     & LM  & Test\\
	  \hline
	  Kaldi (chain) \cite{Povey2012KALDI} \tnote{$\star$} & tri-phone & 3-gram  & 7.43\%  \\
	  Kaldi (nnet3) \cite{Povey2012KALDI} \tnote{$\star$} & tri-phone & 3-gram  & 8.64\%  \\
      ESPnet \cite{Watanabe2018ESPnet}  & mono-char & RNNLM  & 8.0\%  \\
      attention \cite{ACSAttention}  & mono-char & RNNLM  & 18.7\%  \\
      attention \cite{Component-Fusion} & mono-char & RNNLM & 8.71\% \\
	  \hline

	  CAT & mono-phone & 3-gram & 6.34\% \\
	  \hline
	  \end{tabular}
	  \begin{tablenotes}
		\item[$\star$] Using speaker adaptation with i-vectors
	  \end{tablenotes}
	  \end{threeparttable}
	  }
	\vspace{-0.2cm}
	\caption{Aishell results}
	\label{tab:aishell}
	\vspace{-0.2cm}
\end{table}

The WER results on Switchboard are shown in Table 3. Eval2000 consisting of Swichboard evaluation dataset (SW) and Callhome evaluation dataset (CH) is used for evaluation. On Switchboard, our mono-phone system achieves a close performance to bi-phone EE-LF-MMI  system and the results are significantly better than other end-to-end models and mono-phone hybrid LF-MMI system. Compared with bi-phone hybrid LF-MMI, our model is only slightly worse (less than 5\%). The experiment also shows that the performance of our model in CAT can be further enhanced by techniques such as speaker adaptation and RNNLM rescoring.
\begin{table}[tbp]
	\centering
	  \scalebox{0.88}{
	  \begin{threeparttable}
	  \begin{tabular}{|ccc|cc|}
	  \hline
	  Model    & Unit     & LM  & SW & CH \\
	  \hline
	  EE-LF-MMI \cite{SS-LF-MMI} & mono-phone & 4-gram  & 11.0\% & 20.7\% \\
	  EE-LF-MMI  \cite{SS-LF-MMI} & bi-phone & 4-gram  & 9.8\% & 19.3\%  \\
      LF-MMI \cite{SS-LF-MMI} & mono-phone & 4-gram  & 10.7\% & 20.3\% \\
      LF-MMI \cite{SS-LF-MMI} & bi-phone & 4-gram  & 9.5\% & 18.6\%  \\
	  Eesen \cite{Miao2015EESEN}  & mono-phone & 3-gram  & 14.8\% & 26.0\%  \\
      Attention \cite{Zeyer2018RETURNN} & subword &No LM & 13.5\% & 27.1\%  \\
      Seq2Seq \cite{zeyer2018improved} & subword & LSTM  & 11.8\%  & 25.7\% \\
	  \hline

	  CAT & mono-phone & 4-gram & 10.0\% & 19.2\% \\
	  CAT \tnote{$\dagger$}& mono-phone & 4-gram & 9.7\% & 18.8\% \\
	  CAT \tnote{$\dagger$}& mono-phone & RNNLM & 8.8\% & 17.4\% \\
	  \hline
	  \end{tabular}
	  \begin{tablenotes}
		\item[$\dagger$] The experiment uses speaker adaptation with i-vectors so the training is no longer end-to-end.
	  \end{tablenotes}
	  \end{threeparttable}
	  }
	\vspace{-0.2cm}
	\caption{Switchboard results}
	\label{tab:swbd}
\end{table}

We observe that the performance of our model can be further improved simply by increasing the number of the layers (6 to 7, e.g.) or the hidden units per layer (320 to 512, e.g.). Here we only list the results of the neural networks with manageable parameter sizes.

\begin{table}[tbp]
	\centering
	  \scalebox{0.88}{
	  \begin{threeparttable}
	  \begin{tabular}{|c|cc|c|}
	  \hline
	  Dataset   & Model      & Future context & Result\\
	  \hline
   \multirow{4}{1cm}{\centering{WSJ}} & LSTM  & 0 &6.78\\
           & BLSTM & unlimited & 5.9\\
           & VGG-LSTM  & 4 & 6.27 \\
           & TDNN-LSTM  & 7 &  6.42\\
           \hline
       \multirow{4}{1cm}{\centering{Aishell}} & LSTM  & 0 & 8.99\\
           & BLSTM  & unlimited & 7.30\\
           & VGG-LSTM  & 4 & 7.43\\
           & TDNN-LSTM  & 7 & 8.16 \\          
       \hline
   \multirow{4}{1cm}{\centering{Switch\\board}} & LSTM & 0 & 17.7\\
           & BLSTM  & unlimited & 14.6\\
           & VGG-LSTM  & 4 & 17.0\\
           & TDNN-LSTM  & 13 &  16.4\\
           \hline

	  \end{tabular}

	  \end{threeparttable}
	  }
	\vspace{-0.2cm}
	\caption{Latency control experiment results.The experiment is evaluated on dev93 (WER) for WSJ, Aishell test set (CER) and Eval2000 (WER) for Switchboard.}
	\label{tab:aishell}
	\vspace{-0.2cm}
\end{table}
\subsection{Experiment for latency control models}
The results for latency control models are shown in Table 4. The future context is measured by the number of future frames (after sampling) used at current frame. As we use 3*3 convolution in the VGG block and $\{$-1,0,1$\}$ layer-wise context in the TDNN layer, the future context actually depends on how many convolution or TDNN layers we use. We find that the use of VGG blocks and TDNN can alleviate the performance degradation caused by changing the bidirectional model into unidirectional models, without significant increase of the model parameters. 
Specifically, our unidirectional model is comparable with EE-LF-MMI on WSJ and Kaldi-chain model on Aishell. On Switchboard, our unidirectional model (16.4\% WER) is weaker than EE-LF-MMI, but still far better than other end-to-end models (with offline birectional networks) and the online bidirectional network proposed in \cite{onlineNey} (17.3\% WER on Eval2000).
\section{Conclusion}
This paper introduces an end-to-end ASR toolkit - CAT, with the main features of CRF based discriminative training (especially with CTC topology), integrating PyTorch for DNN development, a complete workflow with reproducible examples, and superior results.
CAT obtains the the state-of-the-art results among existing end-to-end models on several major benchmarks, and is comparable with the state-of-the-art hybrid systems. 
To show flexibility, we explore latency control models with various DNN architectures, which enable CAT to be used for streaming ASR. 
We hope CAT, especially the CRF-based framework and software, will be of broad interest to the community, and can be further explored and improved, e.g. the implementation of a more general CRF training framework with different topologies, and the application in more ASR tasks.
\bibliographystyle{IEEEbib}
\bibliography{strings,refs}

\end{document}